\documentclass{article}

\usepackage[nonatbib,final]{neurips_2021}

\usepackage[utf8]{inputenc} 
\usepackage[T1]{fontenc}    
\usepackage{hyperref}       
\usepackage{url}            
\usepackage{booktabs}       
\usepackage{amsfonts}       
\usepackage{nicefrac}       
\usepackage{microtype}      
\usepackage{hyperref}

\usepackage[ruled,vlined,linesnumbered]{algorithm2e}
\usepackage{xspace}
\usepackage{booktabs}
\usepackage{colortbl}
\usepackage{mathtools}
\usepackage{amsmath}
\usepackage{amssymb}
\usepackage{amsthm}
\usepackage{amsfonts}
\usepackage{cleveref}
\usepackage[fleqn]{nccmath}
\Crefformat{section}{Section~#2#1#3}
\Crefformat{proposition}{Proposition~#2#1#3}
\Crefformat{equation}{Eq.\,#2#1#3}

\usepackage{bm}
\usepackage{xspace}
\usepackage{enumitem}
\usepackage[dvipsnames]{xcolor}
\usepackage{subcaption}
\usepackage{caption}
\usepackage{wrapfig}
\usepackage{stackengine}
\usepackage{multirow}
\usepackage{array}
\usepackage{pbox}
\usepackage{dcolumn}
\usepackage[final]{pdfpages}

\usepackage{dsfont}
\usepackage{pifont}
\usepackage{marvosym}
\usepackage{makecell}
\usepackage{cellspace}
\usepackage[super]{nth}

\newcommand{\cmark}{\ding{51}}%
\newcommand{\xmark}{\ding{55}}%
\newcolumntype{Y}{>{\centering\arraybackslash}X}

\newlength\savewidth\newcommand\shline{\noalign{\global\savewidth\arrayrulewidth
  \global\arrayrulewidth 1pt}\hline\noalign{\global\arrayrulewidth\savewidth}}

\makeatletter\renewcommand\paragraph{\@startsection{paragraph}{4}{\z@}
  {.5em \@plus1ex \@minus.2ex}{-.5em}{\normalfont\normalsize\bfseries}}\makeatother
\definecolor{pearDark}{HTML}{2980B9}
\definecolor{pearDarker}{HTML}{1D2DEC}

\newcommand{\HRFormerblock}[6]{
\multirow{3}{*}{
$
\left[
\begin{array}{c}
\text{LSA}, #6, #4 \\
\text{FFN-DW}, #5
\end{array}
\right]
\times #2 \times #3
$
}
}

\newcommand{\convblock}[4]{
\multirow{3}{*}{
$
\left[
\begin{array}{c}
1\times 1, #2 \\
[-.1em] 3\times 3, #2 \\
[-.1em] 1\times 1, #1
\end{array}
\right]
\times #3\times #4
$
}
}

\newcommand{\vecW}{\mathbf{W}}
\newcommand{\vecX}{\mathbf{X}}

\newcommand{\removed}[1]{}

\newcommand{\R}{\mathbb{R}}

\def\eqref#1{equation~\ref{#1}}

\DeclareMathAlphabet{\mathsfit}{\encodingdefault}{\sfdefault}{m}{sl}
\SetMathAlphabet{\mathsfit}{bold}{\encodingdefault}{\sfdefault}{bx}{n}

\def \deitbasedisup {DeiT-B\alambic} 
\def \alambic {\includegraphics[width=0.02\linewidth]{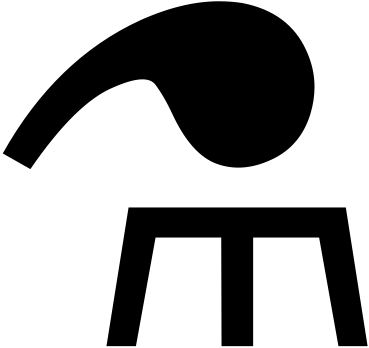}\xspace}

\title{HRFormer: High-Resolution Transformer for Dense Prediction}

\author{
Yuhui Yuan\textsuperscript{124}
\enskip \enskip 
Rao Fu\textsuperscript{1}
\enskip \enskip
Lang Huang\textsuperscript{3}
\enskip \enskip 
\textbf{Weihong Lin\textsuperscript{4}}
\enskip \enskip 
\textbf{Chao Zhang\textsuperscript{3}}
\enskip \enskip 
\textbf{Xilin Chen\textsuperscript{12}}
\enskip \enskip \\
\textbf{Jingdong Wang\textsuperscript{5}\thanks{Corresponding author.}}
\\
\textsuperscript{1}University of Chinese Academy of Sciences
\enskip\enskip\enskip
\textsuperscript{2}Institute of Computing Technology, CAS \\
\enskip\enskip\enskip
\textsuperscript{3}Peking University
\enskip\enskip\enskip
\textsuperscript{4}Microsoft Research Asia
\enskip\enskip\enskip
\textsuperscript{5}Baidu
}

\begin{document}

\maketitle

\begin{abstract}
  We present a High-Resolution Transformer (HRFormer) that learns high-resolution representations for dense prediction tasks,
  in contrast to the original
  Vision Transformer that produces low-resolution representations and has high memory and computational cost.
  We take advantage of the multi-resolution parallel design introduced in high-resolution convolutional networks (HRNet~\cite{WangSCJDZLMTWLX19}), along with local-window self-attention that performs self-attention over small non-overlapping image windows~\cite{huang2019interlaced}, for improving the memory and computation efficiency.
  In addition, we introduce a convolution into the FFN to exchange information across the disconnected image windows.
  We demonstrate the effectiveness of the High-Resolution Transformer on both human pose estimation and semantic segmentation tasks,
  e.g., HRFormer outperforms Swin transformer~\cite{liu2021swin} by $1.3$ AP on COCO pose estimation with $50\%$ fewer parameters and $30\%$ fewer FLOPs.
  Code is available at: {\footnotesize\color{blue}{\url{https://github.com/HRNet/HRFormer}}}.
\end{abstract}

\section{Introduction}
Vision Transformer (ViT)~\cite{dosovitskiy2020image} shows promising performance on ImageNet classification tasks.
Many follow-up works boost the classification accuracy through knowledge distillation~\cite{touvron2020deit}, adopting deeper architecture~\cite{touvron2021going}, directly introducing convolution operations~\cite{graham2021levit,wu2021cvt}, redesigning input image tokens~\cite{yuan2021tokens}, and etc.
Besides, some studies attempt to extend the transformer to address
broader vision tasks such as object detection~\cite{carion2020end}, semantic segmentation~\cite{SETR,Ranftl2021}, pose estimation~\cite{yang2020transpose,li2021pose}, video understanding~\cite{zhang2021multiscale,bertasius2021spacetime,neimark2021video}, and so on. This work focuses on the transformer for dense prediction tasks, including pose estimation and semantic segmentation.

Vision Transformer splits an image into a sequence of image patches of size $16 \times 16$, and extracts the feature representation of each image patch. Thus, the output representations of Vision Transformer lose the fine-grained spatial details that are essential for accurate dense predictions.
The Vision Transformer only outputs a single-scale feature representation,
and thus lacks the capability to
handle multi-scale variation.
To mitigate the loss of feature granularity and model the multi-scale variation,
we present High-Resolution Transformer (HRFormer) that contains
richer spatial information and constructs multi-resolution representations
for dense predictions.

The High-Resolution Transformer is built
by following the multi-resolution parallel design
that is adopted in HRNet~\cite{WangSCJDZLMTWLX19}.
First, HRFormer adopts convolution in both the stem and the first stage
as several concurrent studies~\cite{dai2021coatnet,xiao2021early}
also suggest that convolution performs better in the early stages.
Second, HRFormer maintains a high-resolution stream
through the entire process with parallel medium- and low-resolution streams
helping boost high-resolution representations.
With feature maps of different resolutions,
thus HRFormer is capable to model the multi-scale variation.
Third, HRFormer mixes the short-range and long-range attention
via exchanging multi-resolution feature information
with the multi-scale fusion module.

\begin{figure}[t]
  \small
  \centering
  \includegraphics[width=1.\linewidth]{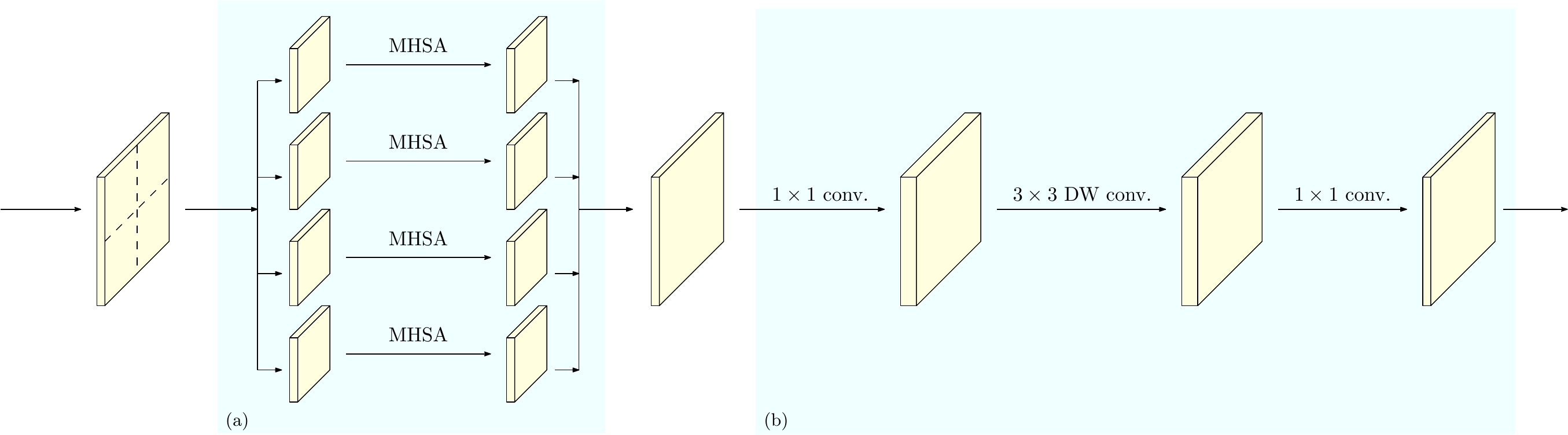}
  \caption{\small \textbf{Illustrating the HRFormer block.}
    The HRFormer block is composed of
    (a) local-window self-attentionm and (b) feed-forward network (FFN) with depth-wise convolution.
    The local-window self-attention scheme is inspired
    by the interlaced sparse self-attention~\cite{yuan2018ocnet,huang2019interlaced}.
  }
  \label{fig:HRFormer_block}
\end{figure}

At each resolution, the local-window self-attention mechanism is adopted to reduce the memory and computation complexity.
We partition the representation maps into
a set of non-overlapping small image windows
and perform self-attention in each image window separately.
This reduces the memory and computation complexity
from quadratic to linear
with respect to spatial size.
We further introduce a $3\times3$ depth-wise  convolution
into the feed-forward network (FFN) that follows the local-window self-attention,
to exchange information
between the image windows
which are disconnected in the local-window self-attention process.
This helps to expand the receptive field
and is essential for dense prediction tasks.
Figure~\ref{fig:HRFormer_block} shows the details of an HRFormer block.

We conduct experiments on image classification, pose estimation, and semantic segmentation tasks, and achieve competitive performance on various benchmarks.
For example,
HRFormer-B gains $+1.0\%$ top-$1$ accuracy on ImageNet classification
over DeiT-B~\cite{touvron2020deit} with $40\%$ fewer parameters and $20\%$ fewer FLOPs.
HRFormer-B gains $0.9\%$ AP over HRNet-W$48$~\cite{XLW19} on COCO $\texttt{val}$ set with with $32\%$ fewer parameters and $19\%$ fewer FLOPs.
HRFormer-B + OCR gains $+1.2\%$ and $+2.0\%$ mIoU over HRNet-W$48$ + OCR~\cite{yuan2019object}
with $25\%$ fewer parameters and slightly more FLOPs
on PASCAL-Context \texttt{test} and COCO-Stuff \texttt{test}, respectively.

\section{Related work}

\paragraph{Vision Transformers.}
With the success of Vision Transformer (ViT)~\cite{dosovitskiy2020image} and the data-efficient image transformer (DeiT)~\cite{touvron2020deit},
various techniques are proposed to improve the ImageNet classification
accuracy of Vision Transformer~\cite{dascoli2021convit,touvron2021going,wu2021cvt,graham2021levit,yuan2021tokens,han2021transformer,chen2021crossvit,liu2021swin,jiang2021token,srinivas2021bottleneck}.
Among the very recent advancements,
the community has verified several effective improvements such as multi-scale feature hierarchies and incorporating convolutions.

For example,
the concurrent works MViT~\cite{fan2021multiscale}, PVT~\cite{wang2021pyramid}, and Swin~\cite{liu2021swin} introduce the multi-scale feature hierarchies into transformer following the spatial configuration of a typical convolutional architecture such as ResNet-$50$.
Different from them, our HRFormer incorporates the multi-scale feature hierarchies through
exploiting the multi-resolution parallel design inspired by HRNet.
CvT~\cite{wu2021cvt}, CeiT~\cite{yuan2021incorporating}, and LocalViT~\cite{li2021localvit}
propose to enhance the locality of transformer via inserting depth-wise  convolutions into either the self-attention or the FFN.
The purpose of the inserted convolution within our HRFormer is different,
apart from enhancing the locality,
it also ensures information exchange
across the non-overlapping windows.

Several previous studies~\cite{ramachandran2019stand,hu2019local} have proposed
similar local self-attention schemes for image classification.
They construct the overlapped local windows following the strided convolution,
resulting in heavy computation cost.
Similar to~\cite{huang2019interlaced,vaswani2021scaling,liu2021swin},
we propose to apply the local-window self-attention scheme
to divide the input feature map into non-overlapping windows.
Then we apply the self-attention within each window independently so as to improve
the efficiency significantly.

There are several concurrently-developed works~\cite{SETR,Ranftl2021} use the
Vision Transformer to address the dense predict tasks such as semantic segmentation.
They have shown that increasing the spatial resolution of the representations
output by the Vision Transformer is important for semantic segmentation.
Our HRFormer provides a different path to address the low-resolution problem
of the Vision Transformer via exploiting the multi-resolution parallel transformer
scheme.

\paragraph{High-Resolution CNN for Dense Prediction.}
The high-resolution convolutional schemes have achieved great success on
both pose estimation and semantic segmentation tasks.
In the development of high-resolution convolutional neural networks, the community has developed three main paths including: (i) applying dilated convolutions to remove some down-sample layers~\cite{chen2017deeplab,yu2015multi},
(ii) recovering high-resolution representations from low-resolution representations with  decoders~\cite{RonebergerFB15,BadrinarayananK17,newell2016stacked,NewellYD16},
and (iii) maintaining high-resolution representations throughout the network~\cite{WangSCJDZLMTWLX19,FourureEFMT017,SaxenaV16,ZhouHZ15,wang2020deep,zhang2020distribution,Huang2020CVPR}.
Our HRFormer belongs to the third path,
and retains the advantages of both
vision transformer and HRNet~\cite{WangSCJDZLMTWLX19}.

\begin{figure}[t]
  \small
  \centering
  \includegraphics[width=1.0\linewidth]{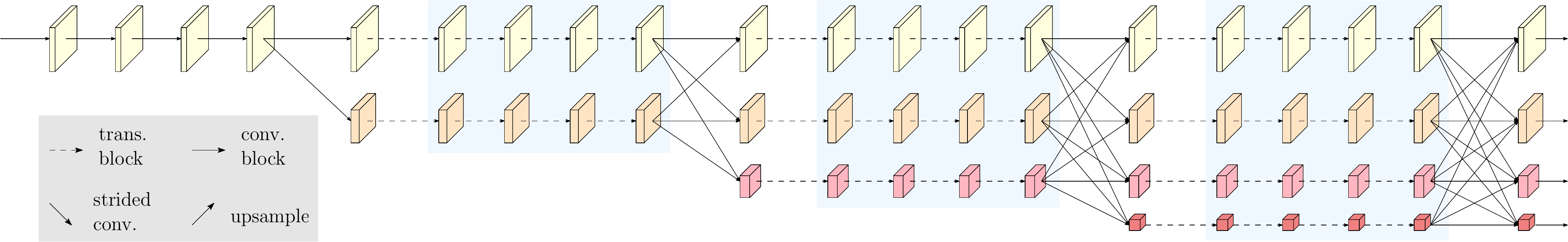}
  \caption{\small \textbf{Illustrating the High-Resolution Transformer architecture.}
    The multi-resolution parallel transformer modules are marked with light blue color areas.
    Each module consists of multiple successive multi-resolution parallel transformer blocks. The first stage is constructed with convolution block and the remained three stages are constructed with transformer block.
  }
  \label{fig:HRFormer_arch}
\end{figure}

\section{High-Resolution Transformer}

\vspace{.1cm}
\noindent\textbf{Multi-resolution parallel transformer.}
We follow the HRNet~\cite{WangSCJDZLMTWLX19} design
and start from a high-resolution convolution stem as the first stage, gradually adding high-to-low resolution streams one by one as new stages. The multi-resolution streams are connected in parallel. The main body
consists of a sequence of stages.
In each stage, the feature representation of each resolution stream is updated
with multiple transformer blocks independently and the information
across resolutions is exchanged repeatedly with the convolutional
multi-scale fusion modules.

Figure~\ref{fig:HRFormer_arch} illustrates the overall HRFormer architecture.
The design of convolutional multi-scale fusion modules exactly follows HRNet.
We illustrate the details of the transformer block in the following discussion
and more details are presented in Figure~\ref{fig:HRFormer_block}.

\vspace{.1cm}
\noindent\textbf{Local-window self-attention.}
We divide the feature maps $\vecX\in \R^{N \times D}$ into
a set of non-overlapping small windows:
$
  \mathbf{X} \rightarrow \{\mathbf{X}_1, \mathbf{X}_2, \cdots, \mathbf{X}_{P}\}
$,
where each window is of size $K \times K$.
We perform multi-head self-attention (MHSA) within each window independently.
The formulation of multi-head self-attention on the $p$-th window is given as:
\begin{align}
  \label{eq:mhsa}
   & {{\rm{MultiHead}}(\vecX_p)} = {\rm{Concat}}[{\rm{head}}(\vecX_p)_1, \cdots, {\rm{head}}(\vecX_p)_H] \in \R^{K^2 \times D},                                                          \\
   & {\rm{head}}(\vecX_p)_{h}  = {\rm{Softmax}}\left[\frac{( \vecX_p\vecW_q^h)( \vecX_p\vecW_k^h)^T }{\sqrt{\nicefrac{D}{H}}} \right] \vecX_p \vecW_v^h \in \R^{K^2 \times \frac{D}{H}}, \\
   & \widehat{\vecX}_p = \vecX_p + {\rm{MultiHead}}(\vecX_p) \vecW_o \in \R^{K^2 \times \frac{D}{H}},
\end{align}
\noindent where $\vecW_o  \in \R^{D \times D}$,
$\vecW_q^h \in \R^{\frac{D}{H} \times D}$,
$\vecW_k^h \in \R^{\frac{D}{H} \times D}$,
and $\vecW_v^h \in \R^{\frac{D}{H} \times D}$
for $h\in\{1,\cdots,H\}$.
$H$ represents the number of heads,
$D$ represents the number of channels,
$N$ represents the input resolutions, and
$\widehat{\vecX}_p$ represents the output representation of MHSA.
We also apply the relative position embedding scheme
introduced in the T$5$ model~\cite{2020t5} to incorporate the relative position
information into the local-window self-attention.

\renewcommand{\arraystretch}{1.1}
\begin{table*}[!htbp]
  \setlength{\tabcolsep}{0.5pt}
  \caption{
    \footnotesize
    \textbf{The architecture configuration of HRFormer.}
    {\textcolor{black}{LSA}}: local-window self-attention, {\textcolor{black}{FFN-DW}}: feed-forward network with a $3\times3$ depth-wise convolution,
    $\left(M_1, M_2, M_3, M_4\right)$: the number of modules,
    $\left(B_1, B_2, B_3, B_4\right)$: the number of blocks,
    $\left(W_1, W_2, W_3, W_4\right)$: the size of windows,
    $\left(H_1, H_2, H_3, H_4\right)$: the number of heads,
    $\left(R_1, R_2, R_3, R_4\right)$: the MLP expansion ratios.
  }
  \label{tab:HRFormer_arch}
  \centering
  \scriptsize
  {
    \begin{tabular}{l|c|c|c|c}
      \shline
      {Res.}                                                             & {Stage $1$}                               & Stage $2$                                                           & Stage $3$                                                           & Stage $4$                                                           \\
      \shline
      \multirow{3}{*}{$4\times$}
                                                                         & \convblock{$256$}{$64$}{$$B_1$$}{$$M_1$$} &
      \HRFormerblock{$C$}{$$B_2$$}{$$M_2$$}{$$H_1$$}{$$R_1$$}{$$W_1$$}   &
      \HRFormerblock{$$C$$}{$$B_3$$}{$$M_3$$}{$$H_1$$}{$$R_1$$}{$$W_1$$} &
      \HRFormerblock{$C$}{$$B_4$$}{$$M_4$$}{$$H_1$$}{$$R_1$$}{$$W_1$$}                                                                                                                                                                                                                                                                 \\
                                                                         &                                           &                                                                     &                                                                     &                                                                     \\
                                                                         &                                           &                                                                     &                                                                     &                                                                     \\
      \hline
      \multirow{3}{*}{$8\times$}
                                                                         &                                           & \HRFormerblock{$$2C$$}{$$B_2$$}{$$M_2$$}{$$H_2$$}{$$R_2$$}{$$W_2$$} & \HRFormerblock{$$2C$$}{$$B_3$$}{$$M_3$$}{$$H_2$$}{$$R_2$$}{$$W_2$$} & \HRFormerblock{$$2C$$}{$$B_4$$}{$$M_4$$}{$$H_2$$}{$$R_2$$}{$$W_2$$} \\
                                                                         &                                           &                                                                     &                                                                     &                                                                     \\
                                                                         &                                           &                                                                     &                                                                     &                                                                     \\
      \hline
      \multirow{3}{*}{$16\times$}
                                                                         &                                           &                                                                     & \HRFormerblock{$4C$}{$$B_3$$}{$$M_3$$}{$$H_3$$}{$$R_3$$}{$$W_3$$}   & \HRFormerblock{$4C$}{$$B_4$$}{$$M_4$$}{$$H_3$$}{$$R_3$$}{$$W_3$$}   \\
                                                                         &                                           &                                                                     &                                                                     &                                                                     \\
                                                                         &                                           &                                                                     &                                                                     &                                                                     \\
      \hline
      \multirow{3}{*}{$32\times$}
                                                                         &                                           &                                                                     &                                                                     & \HRFormerblock{$$8C$$}{$$B_4$$}{$$M_4$$}{$$H_4$$}{$$R_4$$}{$$W_4$$} \\
                                                                         &                                           &                                                                     &                                                                     &                                                                     \\
                                                                         &                                           &                                                                     &                                                                     &                                                                     \\
      \shline
    \end{tabular}
  }
\end{table*}

\renewcommand{\arraystretch}{1.1}
\begin{table}[h]
  \footnotesize
  \centering
  \caption{\footnotesize\textbf{HRFormer instances.}
    HRFormer-T, HRFormer-S, and HRFormer-B represents tiny, small, and base HRFormer model, respectively.
  }
  \label{tab:HRFormer_arch_variants}
  \setlength{\tabcolsep}{8pt}
  \begin{tabular}{l|c|c|c|c}
    \shline
    Model      & \thead{\#modules                                                                                          \\ $\left(M_1, M_2, M_3, M_4\right)$} & \thead{\#blocks\\$\left(B_1, B_2, B_3, B_4\right)$} & \thead{\#channels \\ $\left(C_1, C_2, C_3, C_4\right)$} & \thead{\#heads\\$\left(H_1, H_2, H_3, H_4\right)$} \\
    \shline
    HRFormer-T & $\left(1,1,3,2\right)$ & $\left(2,2,2,2\right)$ & $\left(18,36,72,144\right)$   & $\left(1,2,4,8\right)$  \\
    HRFormer-S & $\left(1,1,4,2\right)$ & $\left(2,2,2,2\right)$ & $\left(32,64,128,256\right)$  & $\left(1,2,4,8\right)$  \\
    HRFormer-B & $\left(1,1,4,2\right)$ & $\left(2,2,2,2\right)$ & $\left(78,156,312,624\right)$ & $\left(2,4,8,16\right)$ \\
    \shline
  \end{tabular}
\end{table}

With \rm{MHSA} aggregates information within each window, we merge them to compute
the output $\mathbf{X}^{\rm{MHSA}}$:
\begin{ceqn}
  \begin{align}
    \{\mathbf{\widehat{X}}_1, \mathbf{\widehat{X}}_2, \cdots, \mathbf{\widehat{X}}_{P}\} \xrightarrow{\rm{Merge}} \mathbf{X}^{\rm{MHSA}}.  \label{eq:merge}
  \end{align}
\end{ceqn}
The left part of Figure~\ref{fig:HRFormer_block} illustrates how local-window self-attention updates the $2$D input representations,
where the multi-head self-attention operates within each window independently.

\vspace{.1cm}
\noindent\textbf{FFN with depth-wise convolution.}
Local-window self-attention performs self-attention
over the non-overlapping windows separately.
There is no information exchange across the windows.
To handle this issue,
we add a $3\times3$ depth-wise convolution in
between the two point-wise MLPs
that form the FFN in Vision transformer:
$\operatorname{MLP} (\operatorname{DW-Conv.}
  (\operatorname{MLP}()))$.
The right part of Figure~\ref{fig:HRFormer_block} shows an example
of how FFN with $3\times3$ depth-wise convolution updates the $2$D input representations.

\vspace{.1cm}
\noindent\textbf{Representation head designs.}
As shown in Figure~\ref{fig:HRFormer_arch},
the output of HRFormer consists of four feature maps of different resolutions.
We illustrate the details of the representation head designs for
different tasks as following:
(i) ImageNet classification,
we send the four-resolution feature maps into a bottleneck and the output channels
are changed to $128$, $256$, $512$, and $1024$ respectively.
Then, we apply the strided convolutions to fuse them and output a
feature map of the lowest resolution with $2048$ channels.
Last, we apply a global average pooling operation followed by the final classifier.
(ii) pose estimation,
we only apply the regression head over the highest resolution feature map.
(iii) semantic segmentation,
we apply the semantic segmentation head over the concatenated representations,
which are computed by first upsampling all the low-resolution representations to the highest resolution and then concatenate them together.

\begin{wrapfigure}{l}{0.5\textwidth}
  \small
  \centering
  \includegraphics[width=0.6\linewidth]{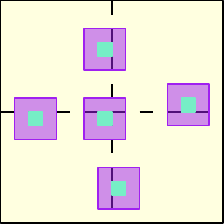}\\
  \caption{\small {Illustrating that FFN with $3\times3$ depth-wise convolution connects the non-overlapping windows.}}
  \label{fig:HRFormer_analysis}
\end{wrapfigure}

\vspace{.1cm}
\noindent\textbf{Instantiation.}
We illustrate the overall architecture configuration of HRFormer
in Table~\ref{tab:HRFormer_arch}.
We use $\left(M_1, M_2, M_3, M_4\right)$ and $\left(B_1, B_2, B_3, B_4\right)$ to represent the
number of modules and the number of blocks of
\{state$1$, stage$2$, stage$3$, stage$4$\}, respectively.
We use $\left(C_1, C_2, C_3, C_4\right)$, $\left(H_1, H_2, H_3, H_4\right)$ and $\left(R_1, R_2, R_3, R_4\right)$ to represent
the number of channels, the number of heads and the MLP expansion ratios in transformer block associated with different resolutions.
We keep the first stage unchanged following the original HRNet
and use the bottleneck as the basic building block.
We apply the transformer blocks in the other stages and each
transformer block consists of a local-window self-attention followed
by an FFN with $3\times3$ depth-wise convolution.
We have not included the convolutional multi-scale fusion modules in Table~\ref{tab:HRFormer_arch}
for simplicity.
In our implementation,
we set the size of the windows on four resolution streams
as $\left(7, 7, 7, 7\right)$ by default.
Table~\ref{tab:HRFormer_arch_variants} illustrates the configuration details of three different HRFormer instances with increasing complexities,
where the MLP expansion ratios $\left(R_1, R_2, R_3, R_4\right)$ are set as $\left(4, 4, 4, 4\right)$ for all models and are not shown.

\vspace{.1cm}
\noindent\textbf{Analysis.}
The benefits of $3\times3$ depth-wise convolution are twofold:
one is enhancing the locality and the other one is enabling the interactions
across windows.
We illustrate how the FFN with depth-wise convolution is capable to expand the
interactions beyond the non-overlapping local windows and model the relations between them in Figure~\ref{fig:HRFormer_analysis}.
Therefore, based on the combination of the local-window self-attention
and the FFN with $3\times3$ depth-wise convolution, we can build the HRFormer block
that improves the memory and computation efficiency significantly.

\section{Experiments}

\subsection{Human Pose Estimation}
\noindent\textbf{Training setting.}
We study the performance of HRFormer
on the COCO~\cite{lin2014microsoft} human pose estimation benchmark,
which contains more than
$200$K images and $250$K person instances labeled with $17$ keypoints.
We train our model on COCO \texttt{train} $2017$ dataset, including $57$K images
and $150$K person instances.
We evaluate our approach on the \texttt{val} $2017$ set and \texttt{test-dev} $2017$,
containing $5$K images and $20$K images, respectively.

We follow most of the default training and evaluation settings
of \texttt{mmpose}~\cite{mmpose2020}\footnote{https://github.com/open-mmlab/mmpose, Apache License 2.0},
and change the optimizer from Adam to AdamW.
For the training batch size, we choose $256$ for HRFormer-T and HRFormer-S
and $128$ for HRFormer-B due to limited GPU memory.
Each HRFormer experiment on COCO pose estimation task takes $8\times$ $32$G-V$100$ GPUs.

\begin{figure*}[t]
  \centering
  \includegraphics[height = 0.171\textwidth]{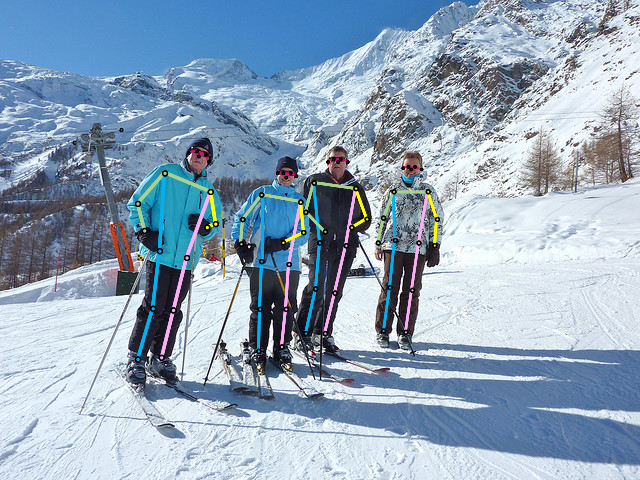}
  \includegraphics[height = 0.171\textwidth]{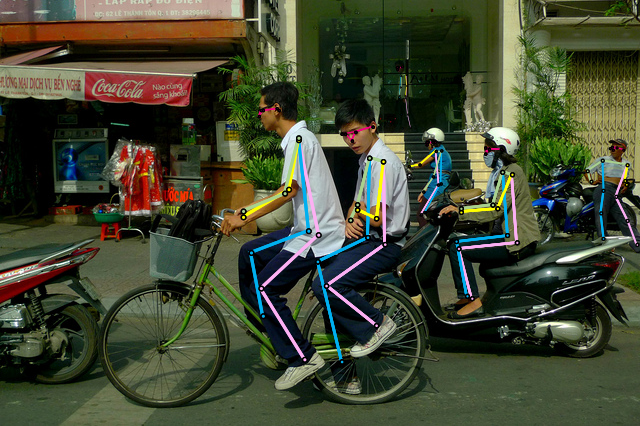}
  \includegraphics[height = 0.171\textwidth]{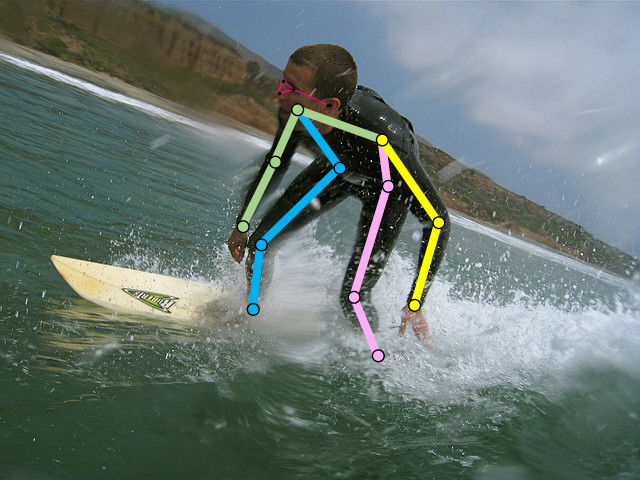}
  \includegraphics[height = 0.171\textwidth]{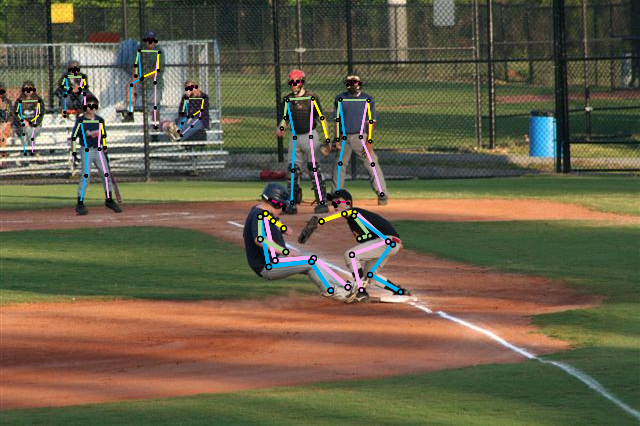}
  \vspace{-0mm}
  \caption{{Example results of HRFormer-B on COCO pose estimation \texttt{val}: containing occlusion, multiple persons, viewpoint and appearance change.}}
  \label{fig:coco_pose_example}
  \vspace{-0mm}
\end{figure*}

\noindent\textbf{Results.}
Table~\ref{tab:coco_pose_val} reports the comparisons on COCO \texttt{val} set.
We compare HRFormer to the representative convolutional method such as HRNet~\cite{XLW19}
and several recent transformer methods, including PRTR~\cite{li2021pose}, TransPose-H-A$6$~\cite{yang2020transpose},
and TokenPose-L/D$24$~\cite{li2021tokenpose}.
HRFormer-B gains $0.9\%$
with $32\%$ fewer parameters and $19\%$ fewer FLOPs when compared to HRNet-W$48$
with an input size of $384\times288$.
Therefore,
our HRFormer-B already achieves $77.2\%$ w/o using any advanced techniques
such as UDP~\cite{Huang2020CVPR} and DARK\cite{zhang2020distribution}. We believe that our HRFormer-B could achieve better results by exploiting either UDP or DARK scheme.
We also report the comparisons on COCO \texttt{test-dev} set in Table~\ref{tab:coco_pose_test}.
Our HRFormer-B outperforms HRNet-W$48$ by around $0.7\%$
with fewer parameters and FLOPs.
Figure~\ref{fig:coco_pose_example} shows some example results
of human pose estimation on COCO \texttt{val} set.

\renewcommand{\arraystretch}{1.1}
\begin{table}[h]
  \footnotesize
  \setlength{\tabcolsep}{5pt}
  \renewcommand{\arraystretch}{1.3}
  \centering
  \caption{\textbf{Comparison on the COCO pose estimation \texttt{val} set.}
    The number of parameters and FLOPs
    for the pose estimation network are measured w/o considering
    neither human detection nor keypoint grouping.
    All results are based on ImageNet pretraining.
    $-$ means the numbers are not provided in the original paper.
  }
  \label{tab:coco_pose_val}
  \footnotesize
  {
    \begin{tabular}{l|ccc|cccccc}
      \shline
      Method                                    & input size               & \#param.                 & FLOPs                   &
      $\operatorname{AP}$                       & $\operatorname{AP}^{50}$ & $\operatorname{AP}^{75}$ & $\operatorname{AP}^{M}$ & $\operatorname{AP}^{L}$ & $\operatorname{AR}$                                             \\\shline
      HRNet-W$32$~\cite{XLW19}                  & $256\times 192$          & $28.5$M                  & $7.1$G                  & ${74.4}$                & ${90.5}$            & ${81.9}$ & ${70.8}$ & ${81.0}$ & ${78.9}$ \\
      HRNet-W$32$~\cite{XLW19}                  & $384\times 288$          & $28.5$M                  & $16.0$G                 & ${75.8}$                & ${90.6}$            & ${82.7}$ & ${71.9}$ & ${82.8}$ & ${81.0}$ \\
      HRNet-W$48$~\cite{XLW19}                  & $256\times 192$          & $63.6$M                  & $14.6$G                 & ${75.1}$                & ${90.6}$            & ${82.2}$ & ${71.5}$ & ${81.8}$ & ${80.4}$ \\
      HRNet-W$48$~\cite{XLW19}                  & $384\times 288$          & $63.6$M                  & $32.9$G                 & ${76.3}$                & ${90.8}$            & ${82.9}$ & ${72.3}$ & ${83.4}$ & ${81.2}$ \\
      PRTR~\cite{li2021pose}                    & $512\times 384$          & $57.2$M                  & $37.8$G                 & ${73.3}$                & ${89.2}$            & ${79.9}$ & ${69.0}$ & ${80.9}$ & ${80.2}$ \\
      TransPose-H-A$6$~\cite{yang2020transpose} & $256\times 192$          & $17.5$M                  & $21.8$G                 & ${75.8}$                & ${-}$               & ${-}$    & ${-}$    & ${-}$    & ${80.8}$ \\
      TokenPose-L/D$24$~\cite{li2021tokenpose}  & $256\times 192$          & $27.5$M                  & $11.0$G                 & ${75.8}$                & ${90.3}$            & ${82.5}$ & ${72.3}$ & ${82.7}$ & ${80.9}$ \\\hline
      HRFormer-T                                & $256\times 192$          & $2.5$M                   & $1.3$G                  & ${70.9}$                & ${89.0}$            & $78.4$   & $67.2$   & ${77.8}$ & ${76.6}$ \\
      HRFormer-T                                & $384\times 288$          & $2.5$M                   & $1.8$G                  & ${72.4}$                & ${89.3}$            & ${79.0}$ & ${68.2}$ & ${79.7}$ & ${77.9}$ \\
      HRFormer-S                                & $256\times 192$          & $7.8$M                   & $2.8$G                  & ${74.0}$                & ${90.2}$            & ${81.2}$ & ${70.4}$ & ${80.7}$ & ${79.4}$ \\
      HRFormer-S                                & $384\times 288$          & $7.8$M                   & $6.2$G                  & ${75.6}$                & ${90.3}$            & ${82.2}$ & ${71.6}$ & ${82.5}$ & ${80.7}$ \\
      HRFormer-B                                & $256\times 192$          & $43.2$M                  & $12.2$G                 & ${75.6}$                & ${90.8}$            & ${82.8}$ & ${71.7}$ & ${82.6}$ & ${80.8}$ \\
      HRFormer-B                                & $384\times 288$          & $43.2$M                  & $26.8$G                 & ${77.2}$                & ${91.0}$            & ${83.6}$ & ${73.2}$ & ${84.2}$ & ${82.0}$ \\
      \shline
    \end{tabular}
  }
\end{table}

\renewcommand{\arraystretch}{1.1}
\begin{table}[h]
  \footnotesize
  \centering
  \setlength{\tabcolsep}{5pt}
  \renewcommand{\arraystretch}{1.3}
  \caption{\textbf{Comparison on the COCO pose estimation \texttt{test-dev} set.}
    The number of parameters and FLOPs
    for the pose estimation network are measured w/o considering
    neither human detection nor keypoint grouping.
    All results are based on ImageNet pretraining.
  }
  \label{tab:coco_pose_test}
  \footnotesize
  {
    \begin{tabular}{l|ccc|cccccc}
      \shline
      Method                                    & input size               & \#param.                 & FLOPs                   &
      $\operatorname{AP}$                       & $\operatorname{AP}^{50}$ & $\operatorname{AP}^{75}$ & $\operatorname{AP}^{M}$ & $\operatorname{AP}^{L}$ & $\operatorname{AR}$                                             \\\shline
      HRNet-W$32$~\cite{XLW19}                  & $384\times 288$          & $28.5$M                  & $16.0$G                 & ${74.9}$                & ${92.5}$            & ${82.8}$ & ${71.3}$ & ${80.9}$ & ${80.1}$ \\
      HRNet-W$48$~\cite{XLW19}                  & $384\times 288$          & $63.6$M                  & $32.9$G                 & ${75.5}$                & ${92.5}$            & ${83.3}$ & ${71.9}$ & ${81.5}$ & ${80.5}$ \\
      PRTR~\cite{li2021pose}                    & $512\times 384$          & $57.2$M                  & $37.8$G                 & ${72.1}$                & ${90.4}$            & ${79.6}$ & ${68.1}$ & ${79.0}$ & ${79.4}$ \\
      TransPose-H-A$6$~\cite{yang2020transpose} & $256\times 192$          & $17.5$M                  & $21.8$G                 & ${75.0}$                & ${92.2}$            & ${82.3}$ & ${71.3}$ & ${81.1}$ & ${-}$    \\
      TokenPose-L/D$24$~\cite{li2021tokenpose}  & $384\times 288$          & $29.8$M                  & $22.1$G                 & ${75.9}$                & ${92.3}$            & ${83.4}$ & ${72.2}$ & ${82.1}$ & ${80.8}$ \\\hline
      HRFormer-S                                & $384\times 288$          & $7.8$M                   & $6.2$G                  & ${74.5}$                & ${92.3}$            & ${82.1}$ & ${70.7}$ & ${80.6}$ & ${79.8}$ \\
      HRFormer-B                                & $384\times 288$          & $43.2$M                  & $26.8$G                 & ${76.2}$                & ${92.7}$            & ${83.8}$ & ${72.5}$ & ${82.3}$ & ${81.2}$ \\
      \shline
    \end{tabular}
  }
\end{table}

\renewcommand{\arraystretch}{1.1}
\begin{table}[h]
  \centering
  \footnotesize
  \setlength{\tabcolsep}{1pt}
  \renewcommand{\arraystretch}{1.3}
  \caption{\textbf{Comparison with the recent SOTA on semantic segmentation tasks.}
    We report the mIoUs on Cityscapes \texttt{val}, PASCAL-Context \texttt{test}, COCO-Stuff \texttt{test}, and ADE$20$K \texttt{val}.
    The number of parameters and FLOPs are measured
    on the image size of $1024\times1024$,
    and the output label map size of $19\times1024\times1024$.
    All results are evaluated with multi-scale testing.
    ${\ddag}$: the results are obtained with extra pre-training on ADE$20$K.
  }
  {
    \begin{tabular}{l|cc|cccc}
      \shline
      Method                                          & \#params. & FLOPs      & {Cityscapes} & {PASCAL-Context} & {COCO-Stuff} & ADE$20$K \\
      \shline
      \multicolumn{6}{l}{\emph{Transformer backbone}}                                                                                      \\\hline
      SETR-PUP~\cite{SETR}                            & $317.8$M  & $2326.7$G  & $82.2$       & $55.3$           & $-$          & $50.1$   \\
      SETR-MLA~\cite{SETR}                            & $309.5$M  & $2138.6$G  & $-$          & $55.8$           & $-$          & $50.3$   \\
      Swin-S + UperNet~\cite{liu2021swin}             & $81.16$M  & $1036.50$G & $-$          & $-$              & $-$          & $49.5$   \\
      Swin-B + UperNet~\cite{liu2021swin}             & $121.18$M & $1187.90$G & $-$          & $-$              & $-$          & $49.7$   \\
      PVT-Large + Semantic FPN~\cite{wang2021pyramid} & $65.1$M   & $-$G       & $-$          & $-$              & $-$          & $43.5$   \\
      \hline
      \multicolumn{6}{l}{\emph{CNN backbone}}                                                                                              \\\hline
      Deeplabv3~\cite{chen2017rethinking}             & $87.1$M   & $1394.0$G  & $80.7$       & $54.1$           & $-$          & $-$      \\
      PSPNet~\cite{zhao2017pyramid}                   & $68.0$M   & $1028.8$G  & $80.0$       & $54.0$           & $43.3$       & $-$      \\
      HRNet-W$48$ + OCR~\cite{yuan2019object}         & $74.5$M   & $924.7$G   & $-$          & $56.2$           & $40.5$       & $45.7$   \\ \hline
      \multicolumn{6}{l}{\emph{CNN+Transformer backbone}}                                                                                  \\\hline
      DPT-Hybrid~\cite{Ranftl2021}                    & $124.0$M  & $1231.5$G  & $-$          & $60.5^{\ddag}$   & $-$          & $49.0$   \\
      HRFormer-B + OCR                                & $56.2$M   & $1119.9$G  & ${82.6}$     & ${58.5}$         & $43.3$       & $50.0$   \\
      HRFormer-B + OCR + SegFix~\cite{yuan2020segfix} & $56.2$M   & $1119.9$G  & ${83.2}$     & $-$              & $-$          & $-$      \\
      \shline
    \end{tabular}
  }
  \label{tab:seg_result}
\end{table}

\begin{figure*}[t]
  \centering
  \includegraphics[height = 0.12\textwidth]{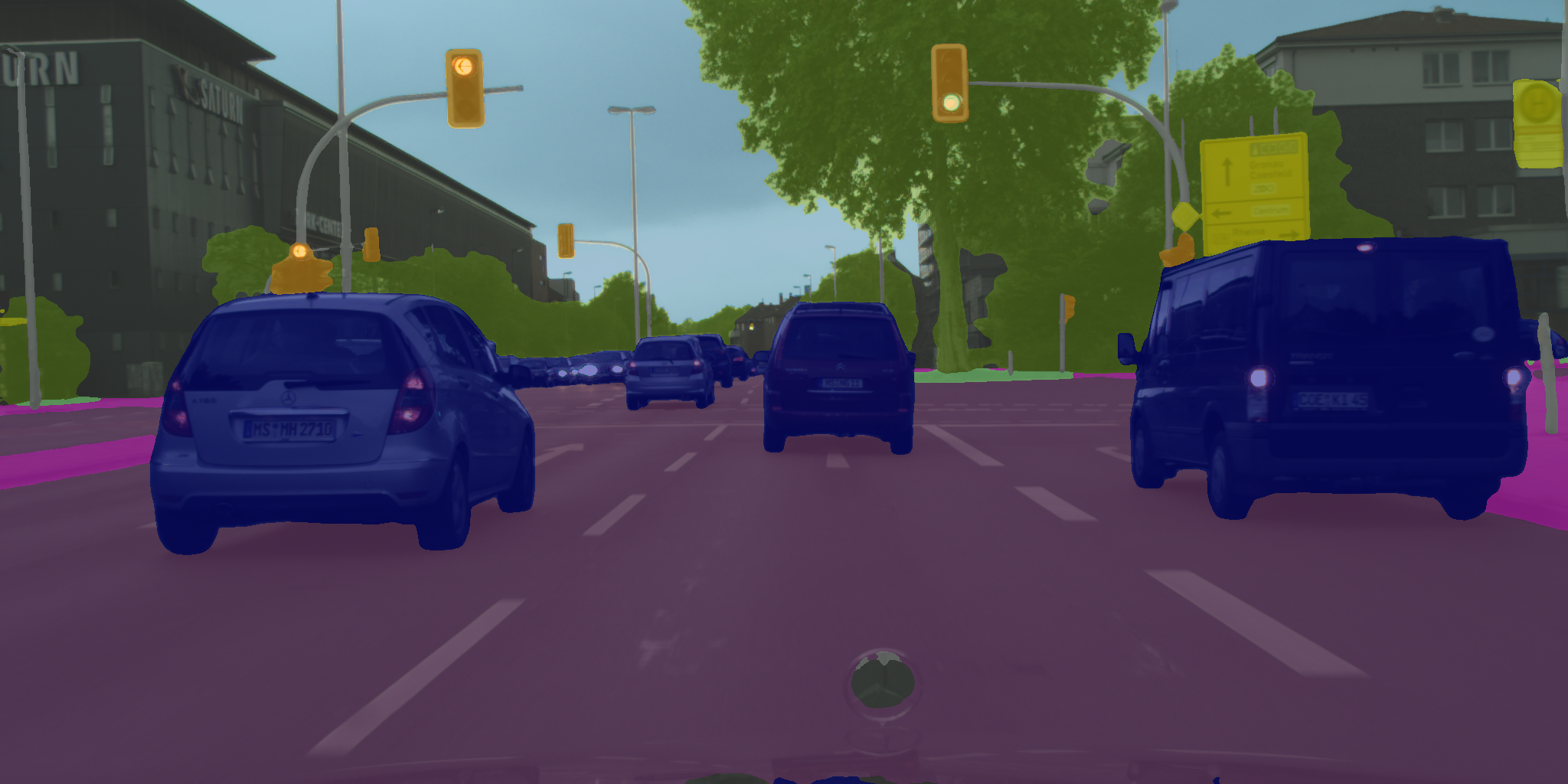}
  \includegraphics[height = 0.12\textwidth]{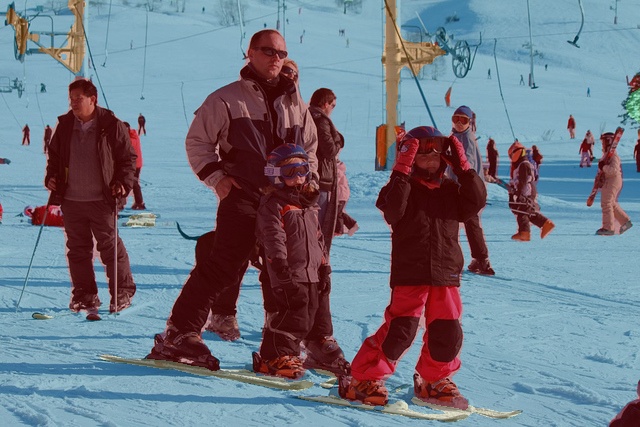}
  \includegraphics[height = 0.12\textwidth]{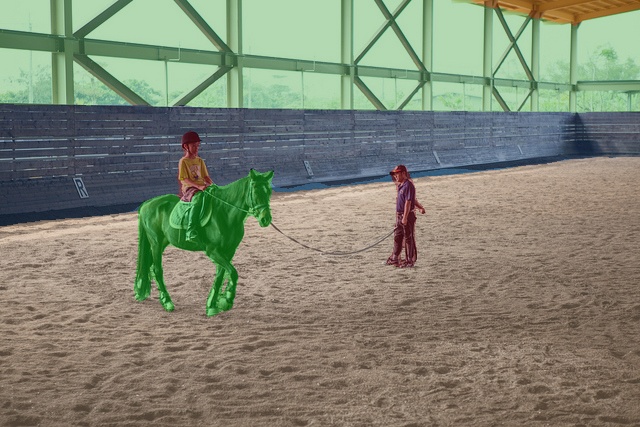}
  \includegraphics[height = 0.12\textwidth]{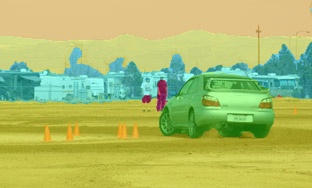}
  \includegraphics[height = 0.12\textwidth]{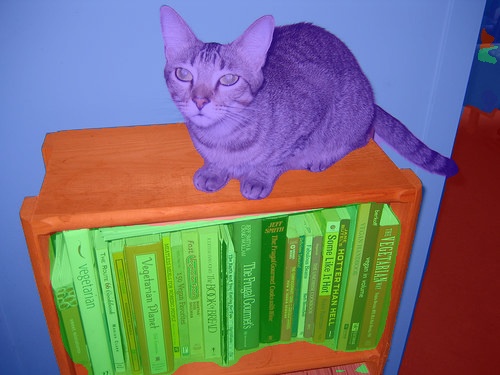}
  \caption{Example results of HRFormer-B + OCR on
      {Cityscapes \texttt{val} (left one), COCO-Stuff \texttt{test} (middle two), and PASCAL-Context \texttt{test} (right two).}}
  \label{fig:seg_result_example}
  \vspace{-.2cm}
\end{figure*}

\subsection{Semantic Segmentation}

\vspace{1mm}
\noindent\textbf{Cityscapes}.
The Cityscapes dataset~\cite{cordts2016cityscapes} is for urban scene understanding.
There are a total of
$30$ classes
and only $19$ classes are used for parsing evaluation.
The dataset contains $5$K high-quality pixel-level finely annotated images and $20$K coarsely annotated images. The finely annotated $5$K images are divided into $2,975$ $\texttt{train}$ images, $500$ $\texttt{val}$ images and $1,525$ $\texttt{test}$ images.
We set the initial learning rate as $0.0001$, weight decay as $0.01$, crop size as $1024\times 512$, batch size as $8$, and training iterations as $80$K by default.
Each HRFormer + OCR experiment on Cityscapes takes $8\times$ $32$G-V$100$ GPUs.

Table~\ref{tab:seg_result} reports the results on Cityscapes \texttt{val}.
We choose to use HRFormer + OCR as our semantic segmentation architecture.
We compare our method with several well-known Vision Transformer based methods~\cite{SETR,Ranftl2021} and CNN based methods~\cite{chen2017deeplab,zhao2017pyramid,yuan2019object}.
Specifically,
SETR-PUP and SETR-MLA use the ViT-Large~\cite{dosovitskiy2020image} as the backbone.
DPT-Hybrid uses the ViT-Hybrid~\cite{dosovitskiy2020image} that consists of a ResNet-$50$ followed by $12$ transformer layers.
Both ViT-Large and ViT-Hybrid are initialized with the weights pre-trained on ImageNet-$21$K,
where both of them achieve around $85.1\%$ top$1$ accuracy on ImageNet.
DeepLabv3~\cite{chen2017deeplab} and PSPNet~\cite{zhao2017pyramid} are based on dilated ResNet-$101$ with output stride $8$.
According to the fourth column of Table~\ref{tab:seg_result},
HRFormer + OCR achieves competitive performance overall.
For example,
HRFormer-B + OCR achieves comparable performance with SETR-PUP while saving $70\%$ parameters
and $50\%$ FLOPs.

\vspace{1mm}
\noindent\textbf{PASCAL-Context}.
The PASCAL-Context dataset~\cite{mottaghi2014role} is a challenging
scene parsing dataset that contains $59$ semantic
classes and $1$ background class.
The \texttt{train} set and \texttt{test} set consist of
$4,998$ and $5,105$ images respectively.
We set the initial learning rate as $0.0001$, weight decay as $0.01$, crop size as $520\times 520$, batch size as $16$, and training iterations as $60$K by default.
We report the comparisons on
the fifth column of Table~\ref{tab:seg_result}.
Accordingly,
HRFormer-B + OCR gains $1.1\%$, $1.5\%$ over HRNet-W$48$ + OCR, SETR-MLA with fewer parameters
and FLOPs, respectively.
Notably, DPT-Hybrid achieves the best performance through extra pre-training the models
on ADE$20$K in advance.
Each HRFormer + OCR experiment on PASCAL-Context takes $8\times$ $32$G-V$100$ GPUs.

\vspace{1mm}
\noindent\textbf{COCO-Stuff}.
The COCO-Stuff dataset~\cite{caesar2018coco} is a challenging
scene parsing dataset that contains $171$ semantic
classes.
The \texttt{train} set and \texttt{test} set consist of
$9$K and $1$K images respectively.
We set the initial learning rate as $0.0001$, weight decay as $0.01$, crop size as $520\times 520$, batch size as $16$, and training iterations as $60$K by default.
We report the comparisons on the last column of Table~\ref{tab:seg_result}
and HRFormer-B + OCR outperforms the previous best-performing HRNet-W$48$ + OCR by nearly $2\%$.
Each HRFormer + OCR experiment on COCO-Stuff takes $8\times$ $32$G-V$100$ GPUs.
Figure~\ref{fig:seg_result_example} shows some example results on Cityscapes,
PASCAL-Context, and COCO-Stuff.

\subsection{ImageNet Classification}
\noindent\textbf{Training setting.}
We conduct the comparisons on ImageNet-$1$K,
which consists of $1.28$M \texttt{train} images and $50$K \texttt{val} images
with $1000$ classes.
We train all models with batch size $1024$ for $300$ epochs with AdamW~\cite{loshchilov2017decoupled} optimizer,
cosine decay learning rate schedule, weight decay as $0.05$,
and a bag of augmentation policies, including rand augmentation~\cite{cubuk2019randaugment},
mixup~\cite{zhang2017mixup}, cutmix~\cite{yun2019cutmix}, and so on.
HRFormer-T and HRFormer-S require $8\times32$G-V$100$ GPUs and HRFormer-B requires $32\times32$G-V$100$ GPUs.

\noindent\textbf{Results.}
We compare HRFormer to some representative CNN methods and vision transformer methods in Table~\ref{tab:imagenet_results}, where all methods are trained on ImageNet-$1$K only.
The results of ViT-Large with larger dataset such as ImageNet-$21$K not included for fairness.
According to Table~\ref{tab:imagenet_results}, HRFormer achieves competitive performance.
For example, HRFormer-B gains $1.0\%$ over DeiT-B
while saving nearly $40\%$ parameters and $20\%$ FLOPs.

\begin{table}[h]
  \footnotesize
  \centering
  \setlength{\tabcolsep}{19 pt}
  \renewcommand{\arraystretch}{1.2}
  \caption{\small\textbf{Comparisons on ImageNet-$1$K \texttt{val}.}
  }
  \label{tab:imagenet_results}
  \begin{tabular}{l|ccc|c}
    \shline
    Method                                        & image size       & \#param. & FLOPs    & {Top-1 acc.} \\
    \shline
    ResNet-$18$~\cite{he2016deep}                 & $224 \times 224$ & $11$M    & $1.8$G   & $69.8$       \\
    ResNet-$50$~\cite{he2016deep}                 & $224 \times 224$ & $26$M    & $4.1$G   & $78.5$       \\
    ResNet-$101$~\cite{he2016deep}                & $224 \times 224$ & $45$M    & $7.9$G   & $79.8$       \\
    \hline
    HRNet-W$18$~\cite{WangSCJDZLMTWLX19}          & $224 \times 224$ & $21.3$M  & $4.0$G   & $76.8$       \\
    HRNet-W$32$~\cite{WangSCJDZLMTWLX19}          & $224 \times 224$ & $41.2$M  & $8.3$G   & $78.5$       \\
    HRNet-W$48$~\cite{WangSCJDZLMTWLX19}          & $224 \times 224$ & $77.5$M  & $16.1$G  & $79.3$       \\ \hline
    RegNetY-$4$G~\cite{radosavovic2020designing}  & $224 \times 224$ & $21$M    & $4.0$G   & $80.0$       \\
    RegNetY-$8$G~\cite{radosavovic2020designing}  & $224 \times 224$ & $39$M    & $8.0$G   & $81.7$       \\
    RegNetY-$16$G~\cite{radosavovic2020designing} & $224 \times 224$ & $84$M    & $16.0$G  & $82.9$       \\
    \hline
    ViT-B/$16$~\cite{dosovitskiy2020image}        & $224 \times 224$ & $86$M    & $55.4$G  & $77.9$       \\
    ViT-L/$16$~\cite{dosovitskiy2020image}        & $224 \times 224$ & $307$M   & $190.7$G & $76.5$       \\
    \hline
    DeiT-T~\cite{touvron2020deit}                 & $224 \times 224$ & $5$M     & $1.3$G   & $72.2$       \\
    DeiT-S~\cite{touvron2020deit}                 & $224 \times 224$ & $22$M    & $4.6$G   & $79.8$       \\
    DeiT-B~\cite{touvron2020deit}                 & $224 \times 224$ & $86$M    & $17.5$G  & $81.8$       \\
    {\deitbasedisup}~\cite{touvron2020deit}       & $384 \times 384$ & $86$M    & $55.4$G  & $83.4$       \\ \hline
    Conformer-T~\cite{peng2021conformer}             & $224 \times 224$ & $23.5$M  & $5.2$G   & $81.3$       \\
    Conformer-S~\cite{peng2021conformer}             & $224 \times 224$ & $37.7$M  & $10.6$G  & $83.4$       \\
    Conformer-B~\cite{peng2021conformer}             & $224 \times 224$ & $83.3$M  & $23.3$G  & $84.1$       \\ \hline
    PVT-T~\cite{wang2021pyramid}                  & $224 \times 224$ & $13.2$M  & $1.9$G   & $75.1$       \\
    PVT-S~\cite{wang2021pyramid}                  & $224 \times 224$ & $24.5$M  & $3.8$G   & $79.8$       \\
    PVT-M~\cite{wang2021pyramid}                  & $224 \times 224$ & $44.2$M  & $6.7$G   & $81.2$       \\
    PVT-L~\cite{wang2021pyramid}                  & $224 \times 224$ & $61.4$M  & $9.8$G   & $81.7$       \\ \hline
    Swin-T~\cite{liu2021swin}                     & $224 \times 224$ & $29$M    & $4.5$G   & $81.3$       \\
    Swin-S~\cite{liu2021swin}                     & $224 \times 224$ & $50$M    & $8.7$G   & $83.0$       \\
    Swin-B~\cite{liu2021swin}                     & $224 \times 224$ & $88$M    & $15.4$G  & ${83.5}$     \\
    Swin-B~\cite{liu2021swin}                     & $384 \times 384$ & $88$M    & $47$G    & $\bf{84.5}$  \\ \hline
    HRFormer-T                                    & $224 \times 224$ & $8.0$M   & $1.8$G   & $78.5$       \\
    HRFormer-S                                    & $224 \times 224$ & $13.5$M  & $3.6$G   & $81.2$       \\
    HRFormer-B                                    & $224 \times 224$ & $50.3$M  & $13.7$G  & $82.8$       \\
    \shline
  \end{tabular}
\end{table}

\subsection{Ablation Experiments}

\begin{table}[h]
  \footnotesize
  \centering
  \setlength{\tabcolsep}{10pt}
  \renewcommand{\arraystretch}{1.3}
  \caption{\label{tab:comp_HRFormer}\textbf{Study
      of the 3$\times$3 depth-wise convolution in FFN.}
    We report the top1 acc., mIoU, and AP on ImageNet \texttt{val}, PASCAL-Context \texttt{test}, and COCO pose estimation \texttt{val}, respectively.
    Results on PASCAL-Context are evaluated with single-scale testing.
    The number of parameters and FLOPs are measured on ImageNet.
  }
  {
    \begin{tabular}{l|cc|ccc}
      \shline
      {Method}                    & \#param. & FLOPs   & {ImageNet} & {PASCAL-Context} & {COCO}  \\
      \shline
      FFN w/o 3$\times$3 DW-Conv. & $7.9$M   & $1.76$G & $77.83$    & $46.84$          & $66.88$ \\
      FFN w/ 3$\times$ 3 DW-Conv. & $8.0$M   & $1.83$G & $78.48$    & $49.74$          & $70.92$ \\
      \shline
    \end{tabular}
  }
\end{table}

\renewcommand{\arraystretch}{1.2}
\begin{table}[h]
  \footnotesize
  \begin{center}
    \setlength{\tabcolsep}{9pt}
    \centering
    \caption{\label{tab:comp_swin}Influence of shifted window scheme \& 3$\times$3 depth-wise convolution within FFN based on Swin-T.}
    \begin{tabular}{l|c|c|c|c}
      \shline
      Method     & 3$\times$ 3 depth-wise convolution in FFN & \#param. & FLOPs  & ImageNet top1 acc. \\
      \shline
      Swin-T     & \xmark                                    & $28.3$M  & $4.5$G & $81.3$             \\
      Swin-T     & \cmark                                    & $28.5$M  & $4.6$G & $82.2$             \\ \hline
      IntraWin-T & \xmark                                    & $28.3$M  & $4.5$G & $80.2$             \\
      IntraWin-T & \cmark                                    & $28.5$M  & $4.6$G & $82.3$             \\
      \shline
    \end{tabular}
  \end{center}
\end{table}

\renewcommand{\arraystretch}{1.2}
\begin{table}[h]
  \footnotesize
  \begin{center}
    \setlength{\tabcolsep}{6pt}
    \centering
    \caption{\label{tab:comp_small_hrnet}Shifted window scheme v.s.
      3$\times$ 3 depth-wise convolution within FFN based on HRFormer-T.}
    \resizebox{1\linewidth}{!}
    {
      \begin{tabular}{c|c|c|c|ccc}
        \shline
        \thead{shifted                                               \\window scheme} & \thead{3$\times$3 depth-wise\\convolution within FFN} & \#param. & FLOPs & \thead{ImageNet \\top1 acc.} & \thead{PASCAL-Context\\mIoU} & \thead{COCO\\AP}  \\
        \shline
        \xmark & \cmark & $8.0$M & $1.8$G & $78.5$ & $49.7$ & $70.9$ \\
        \cmark & \xmark & $7.9$M & $1.6$G & $76.6$ & $43.3$ & $67.3$ \\
        \shline
      \end{tabular}
    }
  \end{center}
\end{table}

\begin{table}[h]
  \footnotesize
  \setlength{\tabcolsep}{24 pt}
  \renewcommand{\arraystretch}{1.2}
  \centering
  \caption{\textbf{Comparisons to ViT \& DeiT on COCO pose estimation \texttt{val}.}
    $\ddag$ marks the methods pretrained on ImageNet-$22$K.
  }
  \label{tab:comp_vit_deit}
  {
    \begin{tabular}{l|ccc|c}
      \shline
      Method            & image size     & \#param. & FLOPs   & {COCO} \\
      \shline
      ViT-Large$^\ddag$ & $256\times192$ & $308.5$M & $60.1$G & $69.2$ \\
      {\deitbasedisup}  & $256\times192$ & $90.0$M  & $17.9$G & $69.0$ \\
      Swin-B$^\ddag$    & $256\times192$ & $93.2$M  & $17.6$G & $74.3$ \\
      \hline
      HRFormer-B        & $256\times192$ & $43.2$M  & $12.2$G & $75.6$ \\
      \shline
    \end{tabular}
  }
\end{table}

\begin{table}[h]
  \footnotesize
  \centering
  \setlength{\tabcolsep}{15 pt}
  \renewcommand{\arraystretch}{1.3}
  \caption{\textbf{Comparisons to HRNet.}
    We report the top1 acc., mIoU, and AP on ImageNet \texttt{val}, PASCAL-Context \texttt{test}, and COCO pose estimation \texttt{val}, respectively.
    Results on PASCAL-Context are based on single-scale testing.
    The number of parameters and FLOPs are measured on ImageNet.
  }
  \label{tab:HRFormer_vs_hrnet}
  {
    \begin{tabular}{l|cc|ccc}
      \shline
      Method     & \#param. & FLOPs   & {ImageNet} & {PASCAL-Context} & {COCO} \\
      \shline
      HRNet-T    & $15.6$M  & $2.7$G  & $76.5$     & $47.8$           & $69.3$ \\
      HRFormer-T & $8.0$M   & $1.8$G  & $78.5$     & $49.3$           & $70.9$ \\  \hline
      HRNet-S    & $24.5$M  & $5.0$G  & $78.7$     & $52.3$           & $73.1$ \\
      HRFormer-S & $13.5$M  & $3.6$G  & $81.2$     & $53.8$           & $74.0$ \\ \hline
      HRNet-B    & $85.3$M  & $20.3$G & $81.4$     & $55.2$           & $75.1$ \\
      HRFormer-B & $50.3$M  & $13.7$G & $82.8$     & $58.5$           & $75.6$ \\
      \shline
    \end{tabular}
  }
\end{table}

\paragraph{Influence of $3\times3$ depth-wise convolution within FFN}
We study the influence of the $3\times3$ depth-wise convolution within FFN
based on HRFormer-T in Table~\ref{tab:comp_HRFormer}.
We observe that applying $3\times3$ depth-wise convolution in FFN significantly
improves the performance on multiple tasks, including ImageNet classification, PASCAL-Context segmentation,
and COCO pose estimation.
For example,
HRFormer-T + FFN w/ 3$\times$ 3 depth-wise convolution
outperforms HRFormer-T + FFN w/o 3$\times$ 3 depth-wise convolution
by $0.65\%$, $2.9\%$ and $4.04\%$
on ImageNet, PASCAL-Context and COCO, respectively.

\paragraph{Influence of shifted window scheme \& 3$\times$3 depth-wise convolution within FFN based on Swin-T.}
We compare our method with the shifted windows scheme of Swin transformer~\cite{liu2021swin} in Table~\ref{tab:comp_swin}.
For fair comparisons, we construct a Intra-Window transformer architecture following the same architecture configurations of Swin-T~\cite{liu2021swin} except that we do not apply shifted windows scheme.
We see that applying 3$\times$3 depth-wise convolution within FFN improves both Swin-T
and Intrawin-T.
Surprisingly, when equipped with 3$\times$ 3 depth-wise convolution within FFN, Intrawin-T even outperforms Swin-T.

\paragraph{Shifted window scheme v.s. 3$\times$3 depth-wise convolution within FFN based on HRFormer-T.}
In Table~\ref{tab:comp_small_hrnet},
we compare the $3\times3$ depth-wise convolution within FFN scheme
to the shifted window scheme based on HRFormer-T.
According to the results, we see that applying 3$\times$3 depth-wise convolution within FFN
significantly outperforms applying shifted window scheme across all different tasks.

\paragraph{Comparison to ViT, DeiT  \& Swin on pose estimation.}
We report the COCO pose estimation results based on the two well-known transformer models, including ViT-Large~\cite{dosovitskiy2020image}, {\deitbasedisup}~\cite{touvron2020deit}
and Swin-B~\cite{liu2021swin} in Table~\ref{tab:comp_vit_deit}.
Notably, both ViT-Large and Swin-B$^\ddag$ are pre-trained on ImageNet$21$K in advance and then finetuned on ImageNet$1$K and achieve $85.1\%$ and $86.4\%$ top-1 accuracy respectively.
  {\deitbasedisup} is trained on ImageNet$1$K for $1000$ epochs and achieves $85.2\%$ top-1 accuracy.
We apply deconvolution modules
to upsample the output representations of the encoder
following the SimpleBaseline~\cite{xiao2018simple} for three methods.
The number of parameters and FLOPs are listed on the fourth and fifth columns of Table~\ref{tab:comp_vit_deit}.
According to the results in Table~\ref{tab:comp_vit_deit},
we see that our HRFormer-B achieves better performance than all three methods with fewer parameters and FLOPs.

\paragraph{Comparison to HRNet.}
We compare our HRFormer to the convolutional HRNet with almost the same architecture configurations
via replacing all the transformer blocks with the conventional basic block consisting
of two $3\times3$ convolutions.
Table~\ref{tab:HRFormer_vs_hrnet} shows the comparison results
on ImageNet, PASCAL-Context, and COCO.
We observe that
HRFormer significantly outperforms HRNet under various configurations
with much less model and computation complexity.
For example, HRFormer-T outperforms HRNet-T by $2.0\%$, $1.5\%$, and $1.6\%$
on three tasks while requiring only around $50\%$ parameters and FLOPs, respectively.
In summary, HRFormer achieves better performance via
exploiting the benefits of transformers such as content-dependent dynamic interactions.

\begin{figure}[t]
  \centering
  \includegraphics[height = 0.24\textwidth]{figs/supp/pose/hrvitw78score_878_id_23359_000000023359.png}
  \includegraphics[height = 0.24\textwidth]{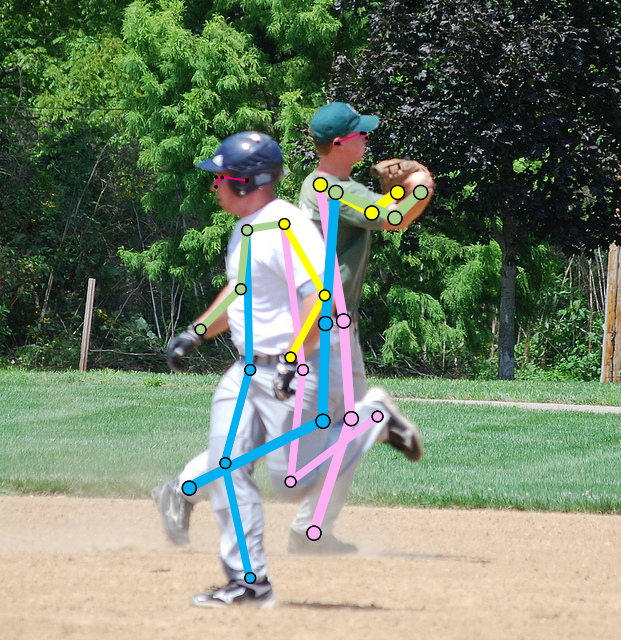}
  \includegraphics[height = 0.24\textwidth]{figs/supp/pose/hrvitw78score_824_id_9772_000000009772.png}
  \includegraphics[height = 0.24\textwidth]{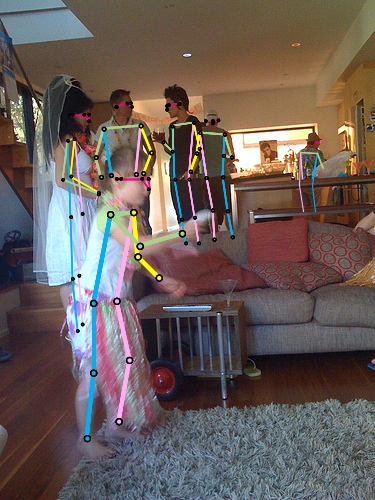}
  \\
  \vspace{1mm}
  \includegraphics[height = 0.235\textwidth]{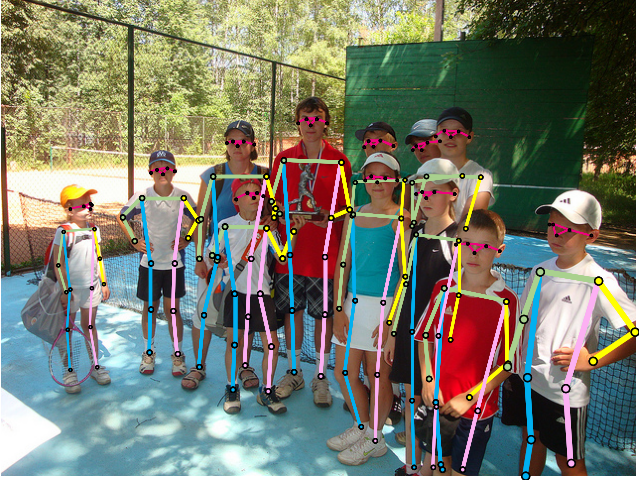}
  \includegraphics[height = 0.235\textwidth]{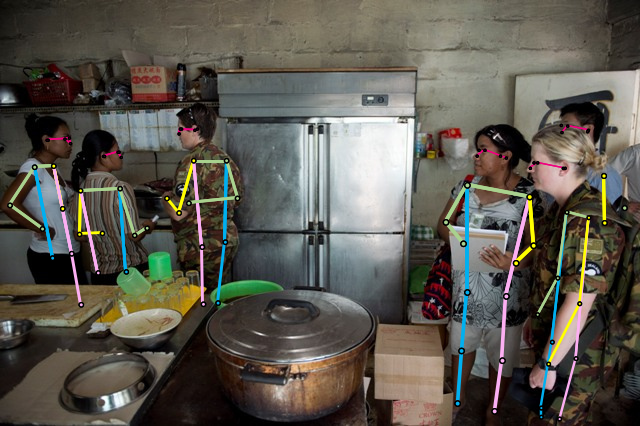}
  \includegraphics[height = 0.235\textwidth]{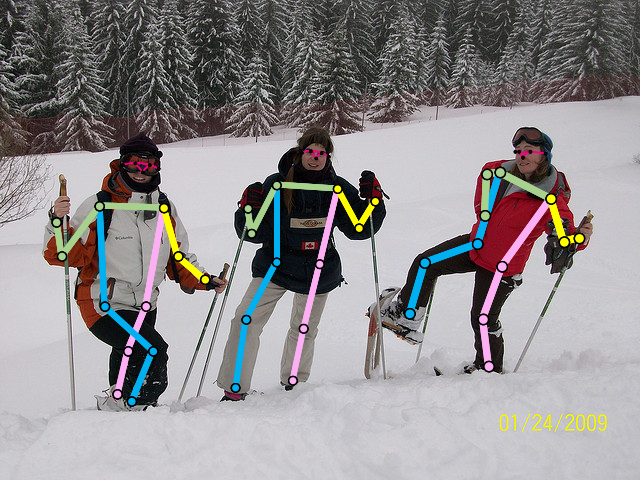}
  \vspace{-0mm}
  \caption{{Visualization of the pose estimation results based on HRFormer-B on COCO  \texttt{val}.}}
  \label{fig:more_pose_example}
  \vspace{-0mm}
\end{figure}

\begin{figure}[hb]
  \centering
  \includegraphics[height = 0.248\textwidth]{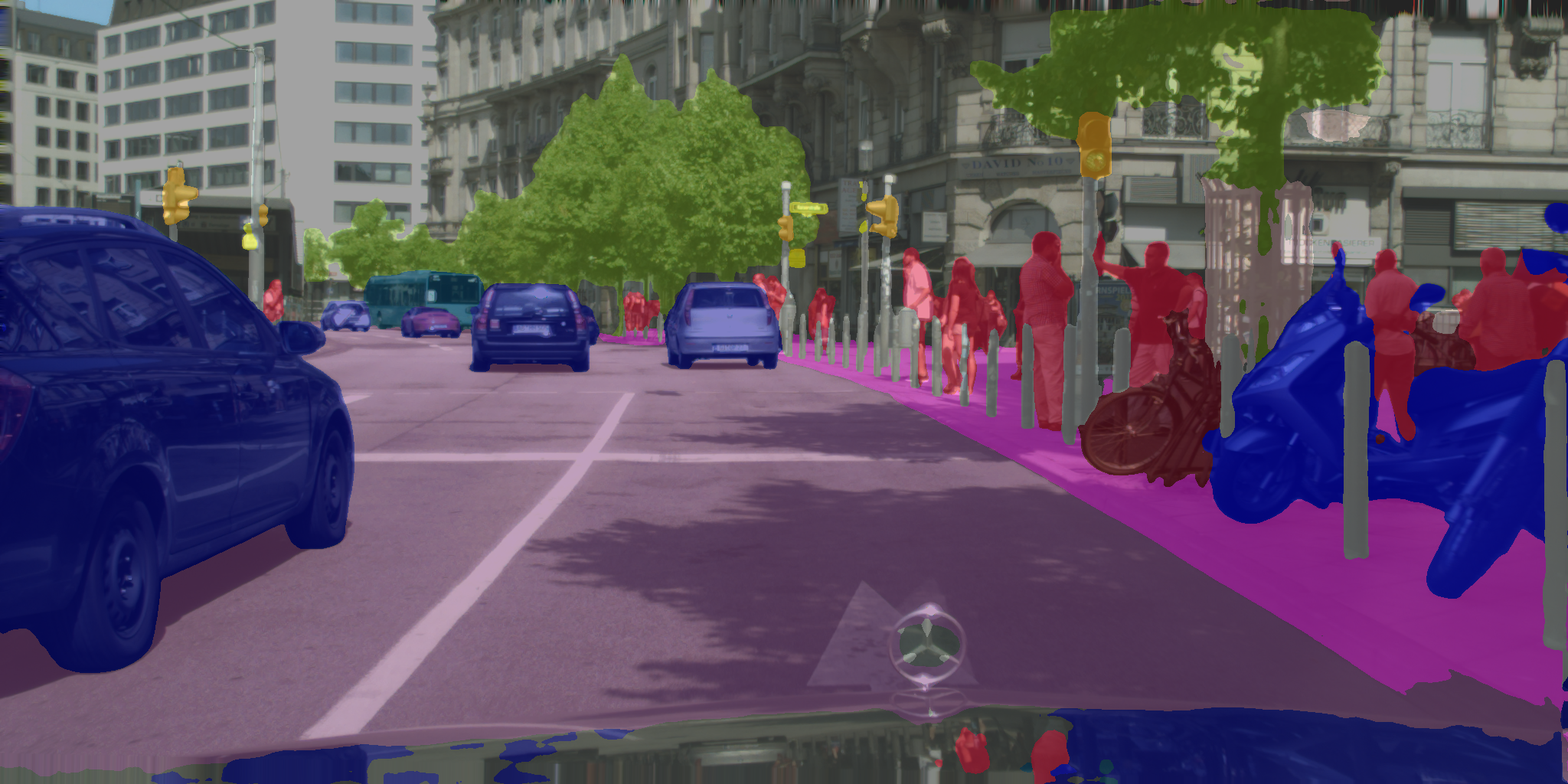}
  \includegraphics[height = 0.248\textwidth]{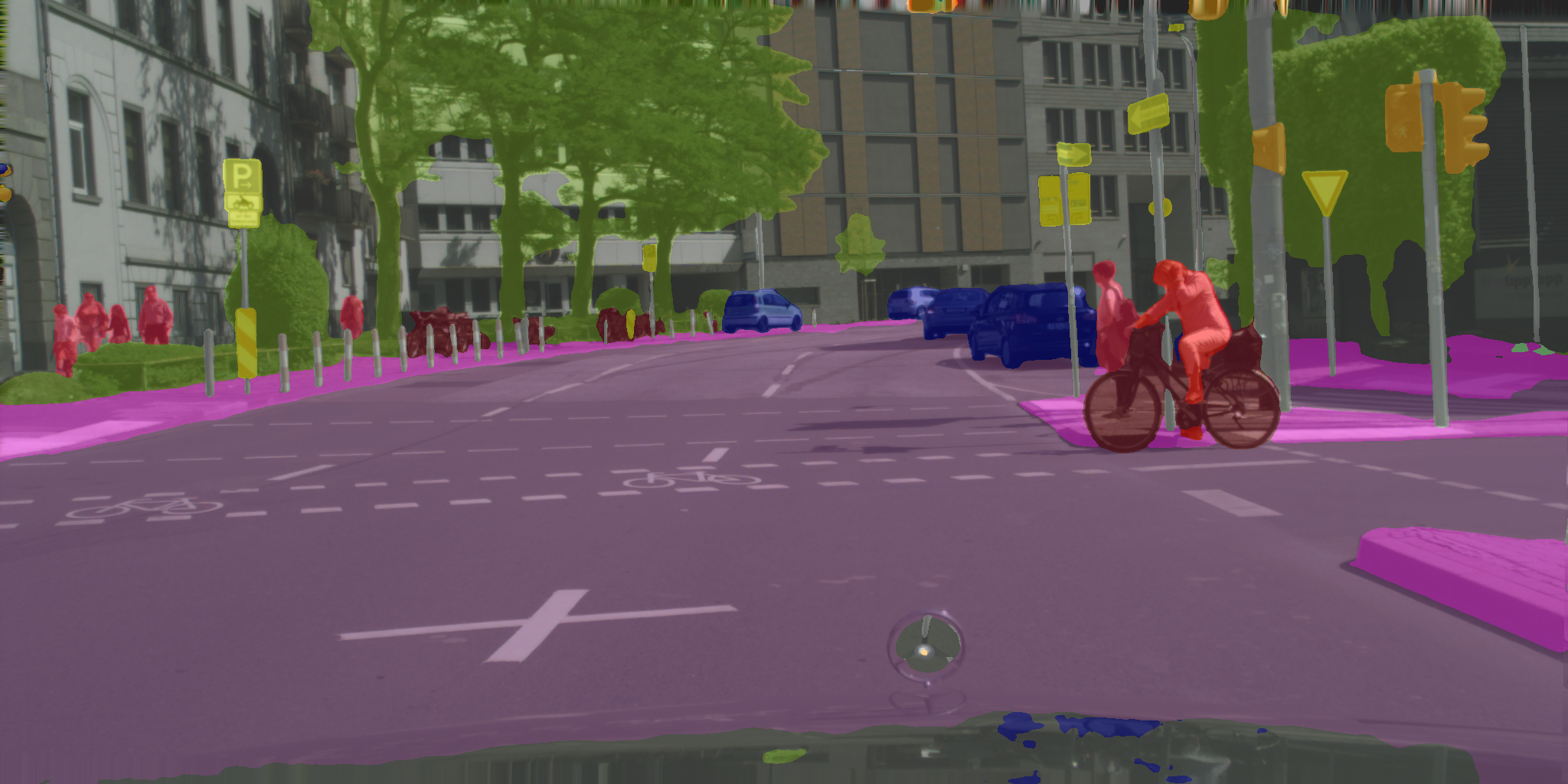}
  \includegraphics[height = 0.245\textwidth]{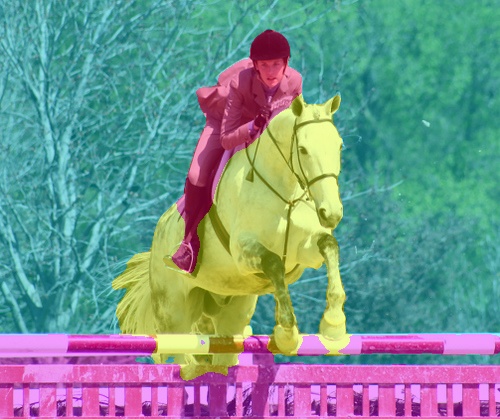}
  \includegraphics[height = 0.245\textwidth]{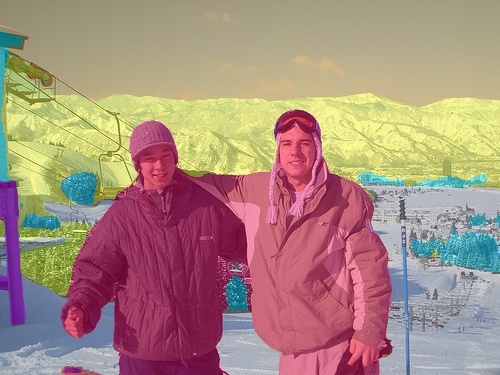}
  \includegraphics[height = 0.245\textwidth]{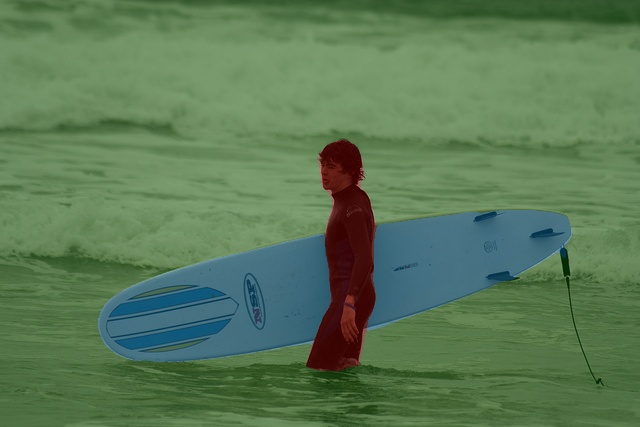}
  \includegraphics[height = 0.247\textwidth]{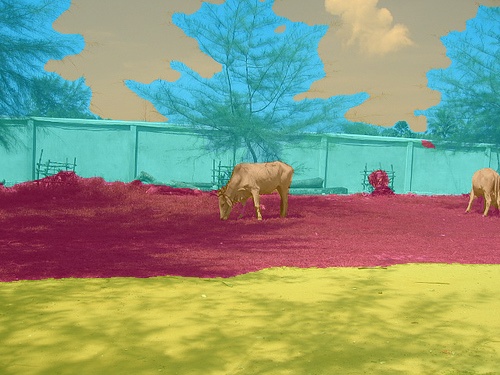}
  \includegraphics[height = 0.247\textwidth]{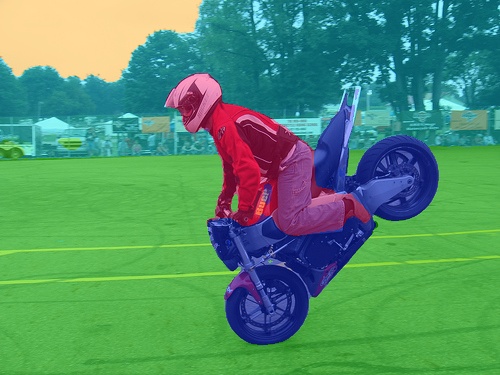}
  \includegraphics[height = 0.247\textwidth]{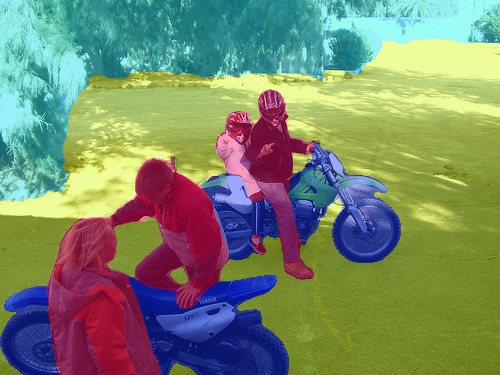}\\
  \caption{{Visualization of the semantic segmentation results based on HRFormer-B + OCR on Cityscapes \texttt{val}, PASCAL-Context \texttt{test}, and COCO-Stuff \texttt{test}.}}
  \label{fig:more_seg_example}
  \vspace{-0mm}
\end{figure}

\section{Conclusion}
In this work, we present the High-Resolution Transformer (HRFormer),
a simple yet effective transformer architecture,
for dense prediction tasks, including pose estimation and semantic segmentation.
The key insight is to integrate the HRFormer block,
which combines local-window self-attention
and FFN with depth-wise convolution to improve the memory and computation efficiency,
with the multi-resolution parallel design of the convolutional HRNet.
Besides, HRFormer also benefits from adopting convolution in the
early stages and mixing short-range and long-range attention
with multi-scale fusion scheme.
We empirically verify the effectiveness of our HRFormer
on both pose estimation and semantic segmentation tasks.

\section{Appendix}

\paragraph{More Visualization Results.}
We present additional visualizations of the example results of our method
on both pose estimation and semantic segmentation tasks.

Figure~\ref{fig:more_pose_example} shows
more pose estimation results of HRFormer-B on COCO \texttt{val}.
Figure~\ref{fig:more_seg_example}
shows more semantic segmentation results on Cityscapes \texttt{val}, PASCAL-Context \texttt{test} and COCO-Stuff \texttt{test}.

\paragraph{Ablation of window sizes.}
We report the results with different window sizes at different resolutions on semantic segmentation tasks and we will add more results if necessary. We use $\left(W_1, W_2, W_3, W_4\right)$ to represent the window sizes associated with feature maps with different resolutions with stride $4$, $8$, $16$, $32$. We choose larger window sizes for higher resolution branches, thus, we have $W_1 > W_2 > W_3 > W_4$. According to these results, we can see that applying larger windows improves the performance, and applying different window sizes at different resolutions makes no big difference.

\begin{table}[!htbp]
  \footnotesize
  \centering
  \setlength{\tabcolsep}{14pt}
  \renewcommand{\arraystretch}{1.2}
  \caption{\label{tab:window_size}\textbf{Influence of the size of windows $\left(W_1, W_2, W_3, W_4\right)$ in HRFormer-B on PASCAL Context.}}
  {
    \begin{tabular}{l|c|cc|c}
      \shline
      Method & {$\left(W_1, W_2, W_3, W_4\right)$} & \#param. & FLOPs   & ss result (ms result) \\
      \shline
      \multirow{8}{*}{HRFormer-B + OCR}
             & $\left(7,7,7,7\right)$              & $56.0$M  & $1051$G & $56.3(57.3)$          \\
             & $\left(9,9,9,9\right)$              & $56.0$M  & $1064$G & $57.4(58.5)$          \\
             & $\left(11,11,11,11\right)$          & $56.1$M  & $1069$G & $56.6(57.6)$          \\
             & $\left(13,13,13,13\right)$          & $56.1$M  & $1083$G & $57.0(58.1)$          \\
             & $\left(15,15,15,15\right)$          & $56.2$M  & $1120$G & $57.5(58.5)$          \\
             & $\left(15,13,11,9\right)$           & $56.1$M  & $1094$G & $56.9(57.9)$          \\
             & $\left(21,17,13,9\right)$           & $56.2$M  & $1148$G & $56.9(57.9)$          \\
             & $\left(17,15,13,11\right)$          & $56.2$M  & $1113$G & $57.5(58.5)$          \\
      \shline
    \end{tabular}
  }
\end{table}

\bibliography{main}
\bibliographystyle{plain}

\end{document}